\documentclass[10pt,twocolumn,letterpaper]{article}

\usepackage{iccv}              
\usepackage{makecell}
\usepackage{multirow}
\usepackage{booktabs}
\usepackage[table]{xcolor}
\usepackage{pifont}
\definecolor{iccvblue}{rgb}{0.21,0.49,0.74}
\usepackage[pagebackref,breaklinks,colorlinks,citecolor=iccvblue]{hyperref}

\author{
Yu Gao$^{1}$\thanks{Equal contribution.} \quad
Anqing Jiang$^{1}$\footnotemark[1] \quad
Yiru Wang$^{1}$\footnotemark[1] \quad
Wang Jijun$^{2}$ \quad
Hao Jiang$^{3}$ \quad
Zhigang Sun$^{1}$ \quad
Heng Yuwen$^{1}$ \\[1pt]
Wang Shuo$^{1}$ \quad
Hao Zhao$^{2}$ \quad
Hao Sun$^{1}$\thanks{Corresponding author: Hao.SUN4@cn.bosch.com.} \\[2pt]
$^{1}$RIX, Bosch \quad $^{2}$AIR, Tsinghua University \quad $^{3}$Shanghai Jiao Tong University \\[1pt]
Team: DiffVLA++
}

\title{DiffVLA++: Bridging Cognitive Reasoning and End-to-End Driving through Metric-Guided Alignment}

\begin{document}
\maketitle
\renewcommand{\thefootnote}{\fnsymbol{footnote}}
\renewcommand{\thefootnote}{\arabic{footnote}}

\begin{abstract}

Conventional end-to-end (E2E) driving models are effective at generating physically plausible trajectories, but often fail to generalize to long-tail scenarios due to the lack of essential world knowledge to understand and reason about surrounding environments. In contrast, Vision-Language-Action (VLA) models leverage world knowledge to handle challenging cases, but their limited 3D reasoning capability can lead to physically infeasible actions. In this work we introduce \textbf{DiffVLA++}, an enhanced autonomous driving framework that explicitly bridges cognitive reasoning and E2E planning through metric-guided alignment. First, we build a VLA module directly generating semantically grounded driving trajectories. Second, we design an E2E module with a dense trajectory vocabulary that ensures physical feasibility. Third, and most critically, we introduce a \emph{metric-guided trajectory scorer} that guides and aligns the outputs of the VLA and E2E modules, thereby integrating their complementary strengths. The experiment on the ICCV 2025 Autonomous Grand Challenge leaderboard shows that our model achieves EPDMS of 49.12.

\end{abstract}

\begin{figure*}[htp]
    \centering
    \includegraphics[width=0.9\textwidth]{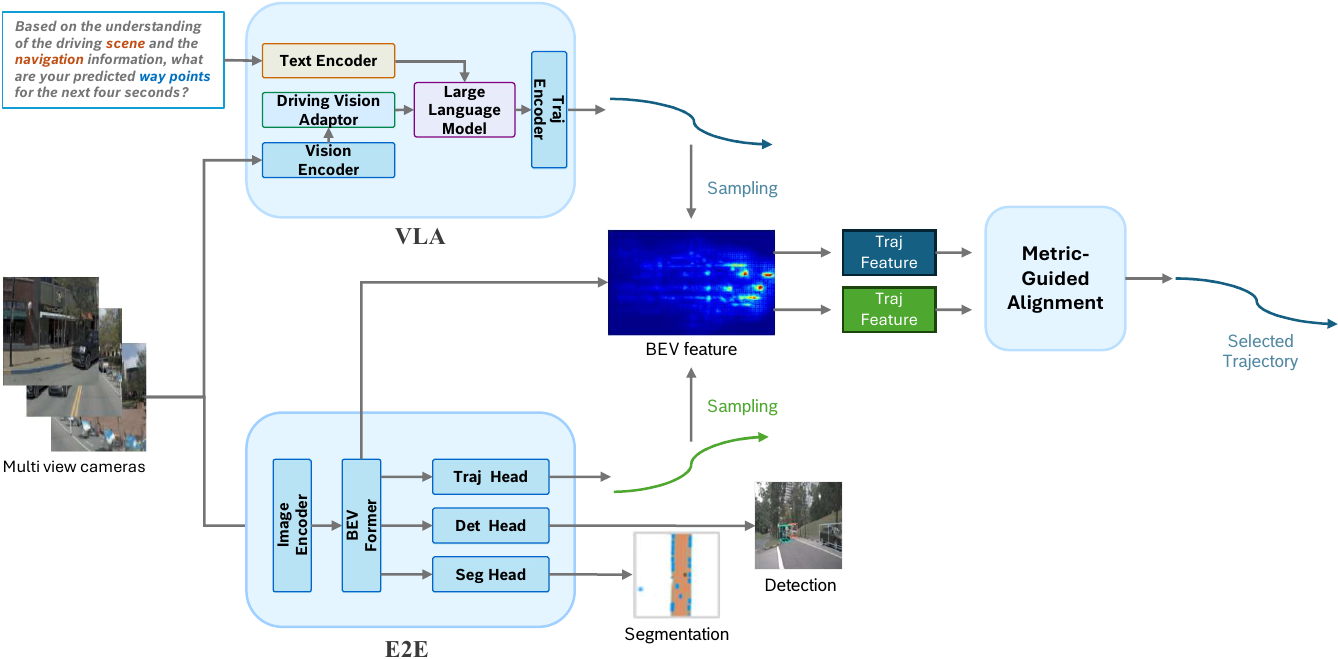}
    \caption{Overview of the DiffVLA++ architecture. It consists of three main components: (i) a VLA module, (ii) a conventional E2E module, both capable of directly generating planning trajectories, and (iii) a trajectory scorer that serves as a metric-guided aligner to unify their outputs.}
    \label{fig:overview}
    \vspace{-10pt}
\end{figure*}

\section{Introduction}
\label{sec:intro}

End-to-end (E2E) autonomous driving frameworks have achieved remarkable progress in recent years by directly mapping multi-modal sensory inputs to control signals or driving trajectories~\cite{hu2022st,chitta2022transfuser,jiang2023vad,tong2023scene,hu2023planning,sun2024sparsedrive,weng2024drive,li2024hydra,li2024enhancing,liao2025diffusiondrive,xing2025goalflow,li2025hydra,zheng2024genad}. These models benefit from powerful spatio-temporal representations and can generate physically plausible driving behaviors under normal conditions. However, they often struggle in long-tail or unseen scenarios due to their limited capability in high-level scene understanding and semantic reasoning~\cite{fu2024drive,zhou2024vision,cai2024driving,chen2024end}. 

DiffVLA~\cite{jiang2025diffvla} attempts to enhance the reliability of spatio-temporal reasoning for both static and dynamic objects by integrating dense and sparse Bird's Eye View (BEV) perception streams. Nevertheless, its limitation arises from the heavy reliance on structured pattern-recognition modules rather than human-level cognitive modeling, which leaves it lacking the generalizable world knowledge that human drivers naturally possess. In DiffVLA++, we retain only the dense BEV branch and aim to address this issue by incorporating cognitive knowledge to enhance reasoning ability.

To further tackle this limitation, recent studies have explored the integration of Large Language Models (LLMs) and Vision-Language-Action (VLA) architectures into autonomous driving systems. Some approaches~\cite{wang2023drivemlm,tian2024drivevlm,jiang2024senna} leverage the extensive world knowledge encoded in LLMs to generate high-level driving decisions, while others~\cite{hwang2024emma,fu2025orion,li2025recogdrive,zhou2025autovla,jiao2025evadrive} directly produce driving trajectories, forming Vision-Language-Action (VLA) models. In DiffVLA, we adopt the former paradigm for its simplicity to integration. In DiffVLA++, we further exploit the VLA framework to generate semantically rich and directly executable driving trajectories.

This raises a key challenge: while E2E models excel at grounded trajectory prediction and VLA models excel at cognitive reasoning with world knowledge, a principled way to bridge these two paradigms has not been fully explored. The difficulty lies in combining semantically rich yet occasionally infeasible VLA trajectories with physically plausible but semantically limited E2E trajectories.

As shown in Fig.~\ref{fig:overview}, we propose \textbf{DiffVLA++}, a framework that explicitly bridges reasoning and planning through a \emph{metric-guided alignment mechanism}. We systematically compare VLA and E2E models on the NavsimV2 benchmark~\cite{cao2025pseudo} and integrate their complementary strengths via the following components:

\begin{itemize}
    \item \textbf{VLA Module}: A fully integrated and differentiable VLA model that generates semantically grounded trajectories with explicit 3D reasoning.  
    \item \textbf{E2E Module}: A dense BEV-based E2E model with a transformer-based trajectory head, ensuring physical feasibility.  
    \item \textbf{Metric-Guided Alignment}: An MLP-based trajectory scorer that shares the BEV feature space with the E2E module, regressing rule-based metrics such as No At-Fault Collisions (NC), Drivable Area Compliance (DAC), and aligning the trajectories from both VLA and E2E to unify their strengths.  
\end{itemize}

Evaluations on the ICCV 2025 Autonomous Grand Challenge benchmark show DiffVLA++ achieves Extended Predictive Driver Model Score (EPDMS)~\cite{cao2025pseudo} of 49.12.

\section{Fully-Differentiable VLA Model for E2E Driving}
\label{sec:pure_vla}

The Vision-Language-Action (VLA) framework is designed around four core modules that jointly enable multimodal reasoning and trajectory generation. The \textbf{visual processing stream} employs a CLIP-based Vision Transformer (ViT-L/14) that encodes multi-view images into a compact set of visual tokens. Each input image is resized and partitioned into patches, which are embedded into high-dimensional representations capturing spatial context. These tokens are then adapted through a Driving Vision Adapter that performs compression and projection, ensuring compatibility with the downstream language model.

In parallel, the \textbf{linguistic stream} encodes navigation commands and high-level driving instructions. A pretrained tokenizer first converts the input into subword units, which are then embedded and processed by a transformer-based text encoder. This produces text tokens that capture both semantic intent and syntactic dependencies.

The two modalities are integrated within a \textbf{large language model} (Vicuna-v1.5-7B), which performs multimodal fusion through a shared embedding space. Visual tokens are concatenated with text tokens, and the model employs causal attention to preserve autoregressive text generation while allowing cross-modal interactions. This design enables the model to jointly reason about visual observations and driving instructions in a unified space.

Finally, the last layer of the LLM projects the fused hidden states into continuous future trajectories of the ego vehicle. Instead of discretizing actions, the model directly predicts waypoints over a four second horizon, where each waypoint includes lateral position, longitudinal position, and heading angle. This fully differentiable design avoids discretization errors, improves smoothness, and allows end-to-end optimization of the driving policy.

To align with the E2E model (Sec.~\ref{sec:e2e}) and the metric-guided trajectory scorer (Sec.~\ref{sec:scorer}), features along the VLA-generated trajectory are sampled from the BEV feature map and fed into the trajectory scorer.

\section{Conventional End-to-End Driving Model}
\label{sec:e2e}

The conventional end-to-end (E2E) driving module in our framework operates on a dense BEV representation generated by BevFormer~\cite{li2024bevformer}. We adopt VoVNet-99~\cite{lee2020centermask} as the image backbone to enhance the visual representational capacity. The BEV feature map is discretized into a $128 \times 128$ grid, covering a spatial extent of $64 \times 64$ meters along the ego-vehicle's $x$ and $y$ axes.

Following the multi-task architecture of Transfuser~\cite{chitta2022transfuser}, the E2E module comprises three prediction heads: 
(i) an \emph{agent detection head} that regresses the states of dynamic agents, 
(ii) a \emph{semantic segmentation head} that predicts scene semantics in BEV space, and 
(iii) a \emph{trajectory planning head} that generates the ego vehicle's future motion.

The agent detection head employs a set of learnable agent queries $\mathbf{Q}_{\text{agent}} \in \mathbb{R}^{N \times d}$ with $N = 32$. Each query interacts with the BEV features via deformable cross-attention to produce agent-centric embeddings $\mathbf{F}_a \in \mathbb{R}^{N \times d}$. These embeddings are decoded into bounding box parameters $(x, y, w, h, \theta)$, representing the center coordinates, width, length, and heading of each detected agent. Together with the semantic segmentation output, this ensures that the BEV representation encodes both dynamic and contextual information for downstream planning (Sec.~\ref{sec:traj_head}) and metric-guided alignment (Sec.~\ref{sec:scorer}).

\subsection{Trajectory Planning Head}
\label{sec:traj_head}
The trajectory planning head operates over a pre-defined dense trajectory vocabulary $\mathcal{V} = \{v_i\}_{i=1}^M$ with $M = 8192$, where each candidate $v_i$ consists of eight waypoints sampled at $2\,\text{Hz}$ over a four second horizon. Each waypoint $\mathbf{p}_t = (x_t, y_t, \theta_t) \in \mathbb{R}^3$ encodes the 2D position and heading angle at time $t$. The vocabulary is constructed via $K$-means clustering on expert trajectories from the navtrain split of the Navsim dataset, ensuring comprehensive coverage of feasible motion patterns.

To encode each candidate trajectory, BEV features are sampled at its waypoints using bilinear grid sampling. This yields a feature sequence for each $v_i$, which is aggregated into a compact embedding through a learned attention mechanism conditioned on the ego state.
Collectively, these operations produce a set of trajectory embeddings $\mathbf{F}_v = [\mathbf{f}_1, \dots, \mathbf{f}_M]^\top \in \mathbb{R}^{M \times d}$, where each $\mathbf{f}_i \in \mathbb{R}^d$ summarizes the visual context along $v_i$.

To incorporate dynamic scene context, $\mathbf{F}_v$ is refined by attending to the agent-centric features $\mathbf{F}_a \in \mathbb{R}^{N \times d}$ from the detection head:
\[
\mathbf{F}_v^{\text{ctx}} = \texttt{CrossAttn}(\mathbf{F}_v, \mathbf{F}_a, \mathbf{F}_a),
\]
where $\texttt{CrossAttn}$ denotes deformable cross-attention, enabling each trajectory query to attend to relevant agent features. The resulting $\mathbf{F}_v^{\text{ctx}} \in \mathbb{R}^{M \times d}$ represents a context-aware set of trajectory hypotheses.

Each refined embedding $\mathbf{f}_i^{\text{ctx}} \in \mathbf{F}_v^{\text{ctx}}$ is then decoded into a residual offset $\Delta v_i \in \mathbb{R}^{8 \times 3}$ via an MLP, yielding the final predicted trajectory:
\[
v_{\text{pred}} = v_i + \Delta v_i,
\]
with $v_i$ selected based on downstream scoring (Sec.~\ref{sec:scorer}). The context-aware embeddings $\mathbf{F}_v^{\text{ctx}}$ are also shared with the metric-guided scorer for joint optimization. $d$ is set to 256 for all modules.

\section{Metric-Guided Alignment}
\label{sec:scorer}

A key novelty of DiffVLA++ lies in \textbf{metric-guided alignment}, which serves as the bridge between the cognitively rich yet occasionally physically inconsistent trajectories from the VLA module and the physically grounded but semantically limited outputs of the E2E module. To achieve this, we employ a lightweight trajectory scorer that maps trajectory features into explicit driving metrics, thereby providing a shared evaluation space for both systems.   

The trajectory scorer is implemented as a set of parallel MLP heads, each regressing one driving metric from the Navsim simulator. Given the context-aware trajectory embeddings $\mathbf{F}_v^{\text{ctx}} = [\mathbf{f}_1^{\text{ctx}}, \dots, \mathbf{f}_M^{\text{ctx}}]^\top \in \mathbb{R}^{M \times d}$ produced by the planning head (Sec.~\ref{sec:traj_head}), the scorer simultaneously predicts metric scores $\hat{s}_m^i$ for each candidate $v_i$ and metric $m$:

\begin{equation*}
\hat{s}_{m}^{i} = \mathrm{MLP}_{m}(\mathbf{f}_i^{\text{ctx}}), \quad 
m \in 
\substack{
\{\text{NC}, \text{DAC}, \text{DDC}, \text{TLC}, \\
\text{EP}, \text{TTC}, \text{LK}, \text{HC}\}
}.
\end{equation*}
Here, the eight driving metrics are categorized as follows:
\begin{itemize}
    \item \textbf{EP}: continuous score in $[0, 1]$ measuring progress along the route centerline.
    \item \textbf{DAC, TLC, TTC, LK, HC}: binary scores in $\{0, 1\}$ indicating compliance with drivable area, traffic lights, collision avoidance, lane keeping, and history comfort.
    \item \textbf{NC, DDC}: ternary scores in $\{0, 0.5, 1\}$ measuring collision and driving direction, where intermediate penalties are assigned when infractions are not directly caused by the ego vehicle.
\end{itemize}

The scorer is trained jointly with the E2E driving model, correlating the BEV feature space with rule-based driving evaluations and providing auxiliary supervision for safer trajectory selection. Ground-truth labels are collected from the navtrain split of the Navsim dataset. Training minimizes a weighted composite objective:
\[
\mathcal{L}_{\hat{s}} = \sum_{i=1}^M \sum_{m} w_m \, \ell_m(\hat{s}_{m}^{i}, s_{m}^{i}),
\]
where $\hat{s}_{m}^{i}$ and $s_{m}^{i}$ denote the predicted and ground-truth scores for metric $m$ and candidate $v_i$, respectively, $w_m$ are per-metric weights balancing scale and importance, and $\ell_m$ is the task-specific loss function:
\begin{itemize}
    \item \textbf{MSE} for continuous metrics (EP);
    \item \textbf{Binary Cross-Entropy (BCE)} for binary metrics (DAC, TLC, TTC, LK, HC);
    \item \textbf{Cross-Entropy} for ternary metrics (NC, DDC).
\end{itemize}


By projecting both VLA-generated and E2E-generated trajectories into this shared metric space, the scorer enables explicit alignment through a common, interpretable performance benchmark.

\section{Post-Processing}
\label{sec:postprocessing}

We first employ a panoptic driving perception model~\cite{wang2025rmt} to predict the drivable area from the front-view camera. Candidate trajectories are projected into this view and discarded if they fall outside the predicted drivable area, serving as a safety check.

After this filtering step, the remaining candidate trajectories are each associated with predicted metric scores. We rank them by computing a weighted sum of the scores:
\begin{equation*}
\begin{aligned}
s_{\text{final}}^{\text{E2E}} &= w_1 \cdot \hat{s}_{\text{NC}} 
                              + w_2 \cdot \hat{s}_{\text{DAC}} 
                              + w_3 \cdot \hat{s}_{\text{EP}} \\ 
                              &\quad 
                              + w_4 \cdot \hat{s}_{\text{TTC}} 
                              + w_5 \cdot \hat{s}_{\text{LK}} 
                              + w_6 \cdot \hat{s}_{\text{DDC}},
\end{aligned}
\end{equation*}
where the weights are empirically set to 
$w_1=4.0$, $w_2=0.8$, $w_3=0.01$, $w_4=0.1$, $w_5=0.04$, and $w_6=6.0$.

We then retain the top-ranked trajectory $traj_{\text{E2E}}$ from the E2E system as its final prediction. Similarly, the trajectory generated by the VLA module, $traj_{\text{VLA}}$, is evaluated using the same trajectory scorer to obtain $s_{\text{final}}^{\text{VLA}}$. Finally, the system selects the overall output trajectory by comparing $s_{\text{final}}^{\text{E2E}}$ and $s_{\text{final}}^{\text{VLA}}$, choosing either $traj_{\text{E2E}}$ or $traj_{\text{VLA}}$ as the final result. Due to time constraints of the competition, the two systems are combined through an offline ensemble.

\vspace{-0.1cm}

\section{Experiments}
\label{sec:experiment}

\subsection{Training for VLA}
\label{sec:training_vla}

For training the VLA model, we adopt the Vicuna-v1.5-7B backbone as the language model and a CLIP ViT-L/14 encoder as the visual backbone. Each input image is resized to $336 \times 336$ and divided into non-overlapping $14 \times 14$ patches, resulting in 196 patches per view. These patches are embedded into 1024-dimensional vectors, producing a sequence of 4096 visual tokens after multi-view concatenation. The Driving Vision Adapter further compresses projects into the joint embedding space of the LLM, producing a compact set of 1024 tokens.

The linguistic input is first tokenized using the LLama tokenizer with a vocabulary size of 32,000. The resulting text tokens are embedded into 1024 text tokens before being fused with visual tokens in the multimodal transformer. The language model has 32 transformer layers, 32 attention heads, and a hidden size of 4096, leading to approximately 7B trainable parameters.

We train the VLA module end-to-end and the model directly regresses future waypoints over a 4-second horizon at 2 Hz, where each waypoint includes $(x, y, \theta)$. Training is performed using AdamW~\cite{loshchilov2017decoupled} with a cosine learning rate schedule~\cite{loshchilov2016sgdr}, an initial learning rate of $1 \times 10^{-5}$. A dropout rate of 0.05 is applied to both the vision adapter and the LLM. The VLA model is trained for one epoch with a batch size of $8$ across eight NVIDIA A800 GPUs.

\subsection{Training for E2E and scorer in Metric-Guided Alignment}

The E2E module and the trajectory scorer are trained jointly to ensure a consistent BEV feature space that captures both semantic and dynamic scene information. The overall training objective includes the following components: agent bounding box regression loss, agent classification loss, semantic segmentation loss, trajectory imitation loss, and scorer loss. We assign loss weights as follows: $1.0$ for agent bounding box regression, $10.0$ for agent classification, $20.0$ for trajectory imitation, $14.0$ for semantic segmentation, and $14.0$ for the scorer. The model is trained for $30$ epochs with a total batch size of $8$ and an initial learning rate of $1 \times 10^{-4}$ on four A800 GPUs. The E2E module use same optimizer and learning rate schdular as VLA module.

\subsection{Experiments result}

We present the performance of the VLA and E2E modules on the Navhard two-stage test in Tab.~\ref{tab:vla_and_e2e_score} and results of final ensembled model in the public leaderboard in Tab.~\ref{tab:metrics_list}

\begin{table}[htbp]
\centering
\caption{Results of different Branches in DiffVLA++ on Navhard Two Stage Test.}
\label{tab:vla_and_e2e_score}
\begin{tabular}{l|c}
\hline
Models & EPDMS \\
\hline
VLA Branch       & 48.0 \\
\hline
E2E Branch      & 43.7 \\
\hline
\end{tabular}
\end{table}

\begin{table}[htbp]
\centering
\caption{Results of DiffVLA++ on the Public Leaderboard}
\label{tab:metrics_list}
\begin{tabular}{l|l}
\hline
\textbf{Metric Name}  & Scores \\
\hline
\textbf{extended\_pdm\_score\_combined}              & \textbf{49.1238} \\
\hline
no\_at\_fault\_collisions\_stage\_one                & 98.2143\\
drivable\_area\_compliance\_stage\_one               & 98.5714\\
driving\_direction\_compliance\_stage\_one           & 100\\
traffic\_light\_compliance\_stage\_one               & 99.2857\\
ego\_progress\_stage\_one                            & 79.5117\\
time\_to\_collision\_within\_bound\_stage\_one       & 98.5714\\
lane\_keeping\_stage\_one                            & 95\\
history\_comfort\_stage\_one                         & 92.8571\\
two\_frame\_extended\_comfort\_stage\_one            & 50 \\
no\_at\_fault\_collisions\_stage\_two                & 88.7709\\
drivable\_area\_compliance\_stage\_two               & 95.3235\\
driving\_direction\_compliance\_stage\_two           & 97.2196\\
traffic\_light\_compliance\_stage\_two               & 98.1711\\
ego\_progress\_stage\_two                            & 73.4289 \\
time\_to\_collision\_within\_bound\_stage\_two       & 87.9888\\
lane\_keeping\_stage\_two                            & 59.4454\\
history\_comfort\_stage\_two                         & 98.9833\\
two\_frame\_extended\_comfort\_stage\_two            & 52.9822\\
\hline
\end{tabular}
\end{table}

\section{Conclusion}
In this work, We propose \textbf{DiffVLA++}, a framework that combines the strengths of VLA and E2E autonomous driving models through metric-guided alignment. By aligning the two systems, our approach achieves an EPDMS of 49.12, surpassing both standalone E2E and VLA models.

\clearpage

\bibliographystyle{IEEEtran}
\bibliography{main}


\end{document}